%% file: ms.tex
\documentclass[10pt,twocolumn,letterpaper]{article}

\usepackage{cvpr}
\usepackage{times}
\usepackage{graphicx}
\usepackage{amsmath}
\usepackage{amssymb}
\usepackage{accents}
\usepackage{subfigure}
\usepackage{stackengine}
\usepackage{enumitem}

\usepackage[latin1]{inputenc}  %% les accents dans le fichier.tex
\usepackage[T1]{fontenc}       %% Pour la césure des mots accentués

\usepackage[hang,flushmargin]{footmisc}
\def\delequal{\mathrel{\ensurestackMath{\stackon[1pt]{=}{\scriptstyle\Delta}}}}


\usepackage[pagebackref=true,breaklinks=true,letterpaper=true,colorlinks,bookmarks=false]{hyperref}

\cvprfinalcopy % *** Uncomment this line for the final submission

\def\cvprPaperID{0012} % *** Enter the CVPR Paper ID here

\ifcvprfinal\fi
\begin{document}

%%%%%%%%% TITLE
\title{Social Ways: Learning Multi-Modal Distributions of Pedestrian Trajectories with GANs}

\author{Javad Amirian\thanks{The research is supported by the CrowdBot H2020 EU Project
		\url{http://crowdbot.org/}}\\
	Univ Rennes, Inria, CNRS, IRISA\\
	France\\
	{\tt\small javad.amirian@inria.fr}
	\and
	Jean-Bernard Hayet\thanks{J.B. Hayet is partially funded by the Intel Probabilistic Computing initiative.}\\
	CIMAT, A.C.\\
	M\'exico\\
	{\tt\small jbhayet@cimat.mx}
	\and
	Julien Pettr{\'e}\\
	Univ Rennes, Inria, CNRS, IRISA\\
	France\\
	{\tt\small julien.pettre@inria.fr}
}

\maketitle

\thispagestyle{empty}
%\todo{1. update fig.2 }\\

%\blfootnote{Research supported by the ERA-Net project ID\_IoT 20CH21\_167534.}

%%%%%%%%% ABSTRACT
\begin{abstract}
This paper proposes a novel approach for predicting the motion of pedestrians interacting with others. It uses a Generative Adversarial Network (GAN) to sample plausible predictions for any agent in the scene. As GANs are very susceptible to mode collapsing and dropping, we show that the recently proposed Info-GAN allows dramatic improvements in multi-modal pedestrian trajectory prediction to avoid these issues. We also left out L2-loss in training the generator, unlike some previous works, because it causes serious mode collapsing though faster convergence.

We show through experiments on real and synthetic data that the proposed method leads to generate more diverse samples and to preserve the modes of the predictive distribution. In particular, to prove this claim, we have designed a toy example dataset of trajectories that can be used to assess the performance of different methods in preserving the predictive distribution modes.
\end{abstract}

%\input{todo} % TODO comment this line

%%%%%%%%% BODY TEXT
\section{Introduction}

Many end-user applications make an intensive use of data analytics about pedestrians motion: urban safety, city planning, marketing, autonomous driving, to name a few ones. Typically, this implies the recollection and the offline analysis of these data, for understanding the pedestrians behaviors and taking decisions about the environment. In some contexts, however, one needs to go further and anticipate, in an online way, what will be the next pedestrian moves and infer their short or mid-term intentions. This allows to trigger early alarms or to take preventive actions when monitoring systems with critical real-time decision-taking processes. In the case of autonomous driving, for example, inferring the intention of the pedestrians surrounding the car is of paramount importance in avoiding collisions. 

\begin{figure}[t]
	\begin{center}
		\includegraphics[width=0.99\linewidth]{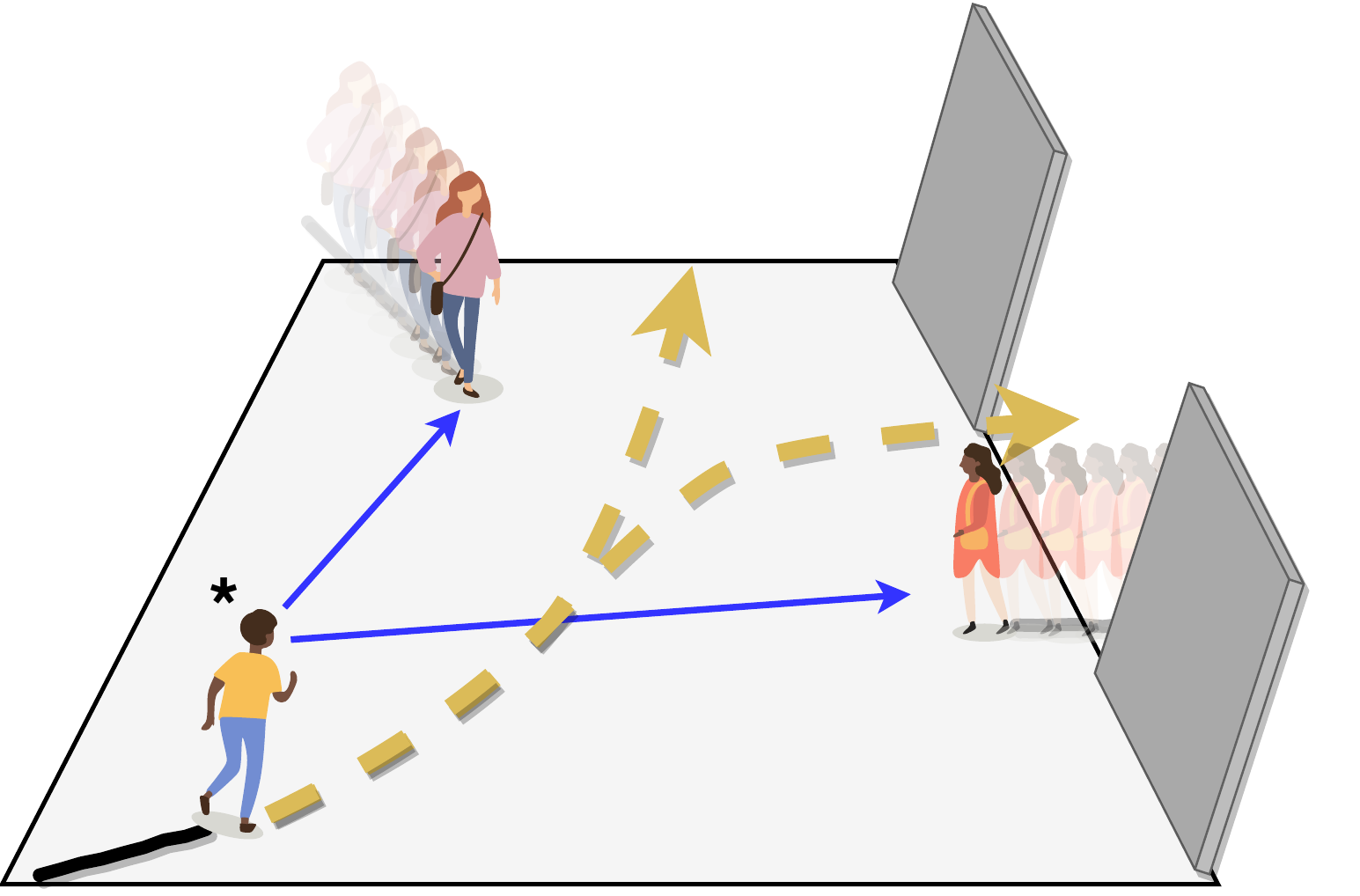}
	\end{center}
	\caption{Illustration of the trajectory prediction problem. Having the observed trajectories of a pedestrian of interest, here shown with a star, and the ones of other pedestrians in the environment, the system should be able to build a predictive distribution of possible trajectories (here with two modes in dashed yellow lines).}
	\label{fig:figOverview}
	\vspace{-0.35cm}
\end{figure}
\begin{figure*}[!ht]
	\begin{center}
		\includegraphics[trim={0cm 2.8cm 0 2.5cm},clip, width=0.95\linewidth]{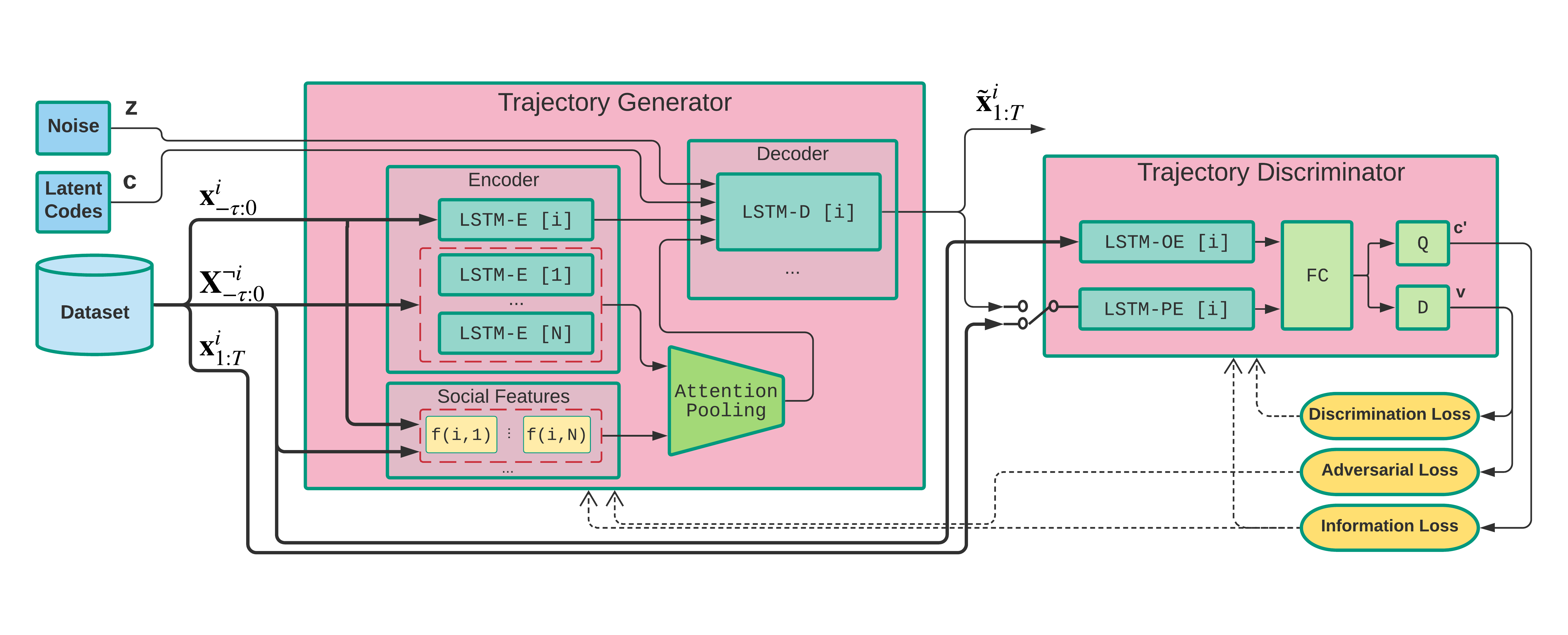}
	\end{center}
	\caption{Block Diagram of the Social Ways prediction system. The yellow ellipses represent loss calculations. The dashed arrows show the backpropagation directions. The bold arrows carry ground truth data.}
	\label{fig:blcok_diagram}
	\vspace{-0.2cm}
\end{figure*}

Nevertheless, this inference problem is extremely complicated to solve. First, because there are many variables which are strongly relevant for the trajectories of single pedestrians: The nature of the surrounding obstacles and their spatial distribution, the nature of the ground, the long-term goal of the pedestrian, his age, his mental state, etc. Then, to make things even more difficult, the motions of a whole set of agents sharing a common space are dependent, through a whole range of interactions that can go from avoidance to meeting intention or person following. A number of interesting studies from neuroscience and bio-mechanics have isolated single factors or optimization principles governing the human motion in very specific contexts (one-to-one interactions, well-stated goals\dots). However, in more general cases, one may rapidly attain the limits of hand-tailored mathematical models. This has motivated the pursuit of more flexible, data-driven statistical approaches that can automatically select the most relevant features for explaining pedestrians walks, and that can benefit from the great efficiency of machine learning techniques.

Our work belongs to the aforementioned category of data-driven methods for predicting the motion of pedestrians in the horizon of a few seconds, given a set of observations of their own past motion and of those of the pedestrians sharing the same space, as illustrated in Fig.~\ref{fig:figOverview}. It relies on a Generative Adversarial Network (GAN)-based trajectory sampler to propose plausible future trajectories. It naturally encompasses the uncertainty and the potential multi-modality of the pedestrian steering decision, which is of critical importance when using this predictive distribution as a belief in higher level decision-making processes.

The main contributions of this work are the following:

\begin{itemize}[leftmargin=*]

\vspace{-0.2cm}
\item An efficient, unsupervised process to train a trajectory prediction GAN architecture based on Info-GAN~\cite{Infogan2016}, without L2 loss, which gives better results than  previous works~\cite{SocialGAN2018,SoPhie2018} in preserving the multi-modal nature of the predictive distribution.

\vspace{-0.2cm}
\item The definition of an attention-based pooling scheme that relies on a few hand-designed interaction features inspired from the neuroscience/bio-mechanics literature, as a form of prior; the best way to combine them to assess the interaction is learned by our system.

\vspace{-0.2cm}
\item The design of a synthetic dataset specifically oriented to the evaluation of the preservation of multi-modality in trajectories predictive distributions.
\vspace{-0.2cm}
\end{itemize}
  
Our architecture is described in Fig.~\ref{fig:blcok_diagram}. It adopts a new strategy to produce plausible samples for an agent from the joint predictive distribution of the set of agents. Our Sampler (Fig.~\ref{fig:blcok_diagram} and Section~\ref{subsec:GAN}) is trained to generate plausible predictions for a single agent, given past observations of trajectories for the whole set of the agents.
%\todo{Note that part of our system relies on a few hand-designed interaction features inspired from the neuroscience/bio-mechanics literature, as a form of prior; the best way to combine them to assess the risk is learned by our system. Also, note that it is possible to remove each of these sub-systems without influencing the others. For example, one could use an energy-based navigation function (like Social Force~\cite{Helbing1995}) instead of our navigation module, or Gaussian Processes to generate samples for the trajectory sampler module.}

\section{Related work}
\label{sec:related}

\paragraph{Closed-form mathematical models.} Many closed-form mathematical models explaining human motion have been introduced in the simulation, graphics and crowd animation areas. Computational geometry-based approaches~\cite{RVO2011} produce optimal motions typically at the limits of collision and not human-like. Optimization-based methods~\cite{Yamaguchi2011} optimize on-the-fly the parameters of an objective function hand-designed to cover relevant aspects of the motion.

In multiple-target tracking, Bayesian techniques typically require prediction processes with simple motion models (random walk or constant velocity) or with parameterized modelling of the social interactions, the goal, etc.~\cite{Pellegrini2009}.

\vspace{-0.4cm}
\paragraph{Data-driven statistical models.} Because of the complexity of pedestrians motion, hand-tailored deterministic models fail to adapt to a wide range of contexts, whereas machine-learning based techniques benefit from large human motion datasets. In~\cite{Leal2014}, for tracking pedestrians from a vehicle, interaction features useful for avoidance are learned from optical flow data. In~\cite{Keller2014}, pedestrian path prediction, in the same context of mobile sensing, is done in a low-dimensional latent space through Gaussian process dynamical models with augmented features extracted from the video optical flow. In~\cite{Trautman2015}, interacting mixtures of Gaussian Processes (GPs) are used for predicting the whereabouts of goal-driven social agents in crowds, where the parameters are learned from training data.  

\vspace{-0.4cm}
\paragraph{NN-based data-driven models.} With the advent of NN-based machine learning, the sequential nature of motion has motivated the use of Recurrent Neural Networks or more efficient variants, such as LSTMs~\cite{Graves2014}, for the prediction task. The Social-LSTM architecture~\cite{SocialLSTM2016} associates each agent to a LSTM network and a social pooling aggregates the hidden states of the neighboring agents, to form an interaction feature. Then, each agent interaction feature is combined with its own hidden state to generate the predicted positions for the future frames, with another LSTM network.

%In~\cite{Li2017}, from crowd video sequences instead of individual trajectories, meaningful global patterns and motion features are extracted with layers of Convolutional LSTM units, which allows prediction about the crowd motion.

In~\cite{SocialAttention2018Vemula}, groups of agents are modeled as spatio-temporal graphs where edges (temporal and spatial) are associated to RNNs. Temporal edges capture the evolution of single humans while spatial edges capture the evolution of agent-to-neighbors relationships. These hidden features are combined linearly to produce an influence score feeding the temporal network. The prediction output takes the form of a bivariate Gaussian distribution. %In~\cite{Hasan2018}, the authors use the same LSTM with social pooling methodology while also integrating head pose estimates, and improve the results for slow pedestrian motions.

In~\cite{CIDNN2018}, a Crowd Interaction Deep Neural Network uses four modules: A trajectory encoding module encodes individual trajectories using LSTMs units; A location encoding module maps the locations of the pedestrians and the influence they have on each other; An interaction module forms linear combinations of other agents trajectory encodings, weighted by their influence; Finally, the predicted trajectory is determined by sending this linear combination through a fully connected layer. The reported results look promising, however we were not able to reproduce them entirely.

In~\cite{Pfeiffer2018}, LSTMs capture the evolution of single trajectories, while the interaction history is handled through a LSTM fed with histograms of closest distances over an angular discretization of the surrounding, while the local obstacles are embedded in an occupancy grid.

\vspace{-0.3cm}
\paragraph{Handling the multimodal nature of predictions with generative NNs.} In many situations, the predictive distribution of a pedestrian motion is inherently multi-modal, e.g., at a crossroads. Without a proper modeling of this multi-modality, RNN-based methods, given observed trajectories with multiple possible outcomes, may simply be condemned to average all the possible outputs. The DESIRE architecture~\cite{DESIRE2017} handles this multi-modality. A Sample Generation Module based on variational auto-encoders generates samples of potential outcome trajectories and the Ranking and Refinement Module evaluates a learned long-term score associated to the sampled trajectories and refines these trajectories, in an inverse optimal control scheme.

In~\cite{PlausiblePaths2017}, a social-aware LSTM, similar to~\cite{SocialLSTM2016}, embeds the prior from the training data as hidden feature. Motion variability is taken into account by using layered Gaussian processes acting on the hidden features of the LSTMs.

Finally, following the success of Generative Adversarial Networks (GAN)~\cite{Goodfellow2014} in other areas to learn data distributions and produce new samples~\cite{BEGAN2017}, Gupta et al. have proposed a trajectory sampler that handles the interactions between all the observed pedestrians by pooling the GAN input random vector with a vector combining the hidden representations of the other pedestrians trajectories~\cite{SocialGAN2018}.

\section{Problem statement and system overview}
\input{approach}

\section{Experimental results}
\input{experiments}
%\input{experiments_temp}

\section{Conclusions and Future Works}

We have presented a novel approach for the prediction of pedestrians trajectories among crowds. It uses an Info-GAN to produce samples from the predictive distribution of individual trajectories, and integrates a few hand-designed interaction features inspired from the neuroscience/bio-mechanics literature, as a form of prior over the attention pooling process. We have shown through extensive evaluations on commonly used datasets that this approach partly improves the prediction accuracy of state-of-the-art methods on the datasets where the predictive distributions have the largest variances. We have also proposed a specifically designed dataset and an evaluation benchmark to show that Info-GANs achieve the best results in preserving multi-modality, compared with other variants. Finally, we are aware that is still room for improving the current generative models in pedestrian motion prediction and, above all, for exploiting these models in decision making.

{\small
\bibliographystyle{ieee}
\bibliography{ms}
}

\end{document}

%% file: approach.tex
% !TEX root = egpaper_for_review.tex

\input{notations}

\subsection{Notations and problem formulation} 

In the following, we use indices $i,j\in \{1, ..., N\}$ to refer to pedestrians, where $N$ is the total number of pedestrians; a single observation of pedestrian $i$ in the scene at time $t$ is denoted by the $4\times 1$ vector $\mathbf{x}^i_t$, which itself contains the position $\mathbf{p}^{i}_t$ and velocity $\mathbf{v}^{i}_t$ of the pedestrian: 
$\mathbf{x}^i_t \delequal ((\mathbf{p}^{i}_t)^T, (\mathbf{v}^{i}_t)^T)^T$.  We assume that we have access to $\tau+1$ 
consecutive observed samples $\mathbf{x}^i_{-\tau: 0}$ of the pedestrians trajectory
for each $i \in \{1, ..., N\}$. 
We also handle the set of observed samples of all pedestrians except $i$ with 
$\mathbf{X}^{\neg i}_{-\tau: 0} \delequal \{ \mathbf{x}^j_{-\tau: 0} | j \in \{1, ..., N\}, j \neq i \}$.

The problem is then to predict the trajectories of each pedestrian for the next $T$ time steps, i.e. $\mathbf{x}^i_{1: T}$. 

The rationale behind our approach is the following: When deciding his steering actions, a pedestrian anticipates likely scenarios about the evolution of his surrounding in the near future. Now, this anticipation may not be always very easy, because of the uncertainties in the neighbors future motion and intentions. In most recent NN-based motion prediction systems~\cite{SocialAttention2018Vemula,CIDNN2018,Pfeiffer2018}, the input is taken as the set of most recent observations of the surrounding pedestrians. Hence, the mappings from observations to predicted trajectories built through the networks do not consider explicitly the uncertain and multimodal nature of the neighbors future trajectories, and, in a way, the network is expected to learn it too, which may be too much to expect.   

\vspace{-0.1cm}
\subsection{GAN-based Individual Trajectory Sampler}
\label{subsec:GAN}

Our Social Ways GAN generates independent random trajectory samples that mimic the distribution of trajectories among our training data, conditioned on observed initial tracklets of duration $\tau$ for all the agents in the scene. This system is depicted in Fig.~\ref{fig:blcok_diagram}. It takes as an input the observed trajectories of $N$ pedestrians, $\mathbf{X}_{-\tau: 0}$ and a random vector $\mathbf{z}$ sampled from a fixed distribution $p_z$. It samples a plausible trajectory $\tilde{\mathbf{x}}^{i,k}_{1: T}$ for agent $i$ for the next $T$ time steps, where $k$ identifies one generated sample. The network should learn the whereabouts of an agent altogether with the impact a surrounding crowd has on its trajectory.

A GAN contains two components that act in opposition to each other during the training phase~\cite{Goodfellow2014}. The Discriminator $D$ is trained to detect fake samples from real ones, while the Generator $G$ should produce new samples that fool the Discriminator and confuse its predictions. In a conditional version, both the Generator and the Discriminator are conditioned on some given data. Here, our GAN is conditioned on recent observations  $\mathbf{x}^i_{-\tau: 0}$, for agent $i$, and  $\mathbf{X}^{\neg i}_{-\tau:0}$, for the other agents, and the Generator uses a noise vector $\mathbf z$ to complete $\mathbf{x}^i_{-\tau: 0}$ into a full trajectory $G(\mathbf z|\mathbf{x}^i_{-\tau: 0},\mathbf{X}^{\neg i}_{-\tau:0})$.

\vspace{-0.3cm}
\subsubsection{Description of the Generator network}

Our system shares a number of characteristics with existing trajectory generation systems~\cite{SocialGAN2018,SoPhie2018} but it also includes critical novelties. The Generator network uses one LSTM layer (denoted as LSTM-E) to learn the temporal features along trajectories.  The encoding of past trajectories $ \mathbf{x}^i_{-\tau:0}$ for an agent is similar to~\cite{SocialGAN2018}. The LSTM-E cell encodes the history of the agent $i$ through the recursive application of:

\vspace{-0.2cm}
\begin{equation}
\begin{array}{ccc}
\mathbf{h}^i_t&=& \lambda^e(\mathbf{h}^i_{t-1},\mu(\mathbf{x}^i_t;\mathbf W_{\mu}); \mathbf W_{\lambda^e})
\end{array}
\end{equation}
with $t\in[-\tau,0]$, $\mu$ a linear embedding of the agent state and $\lambda^e$ the cell of LSTM-E. $\mathbf{h}^i_t$ is the hidden state vector in LSTM-E at time $t$. It is depicted at the left part of Fig.~\ref{fig:blcok_diagram}.

For the decoding process and the generation of samples, we apply a similar process through another LSTM layer (denoted as LSTM-D) with hidden state $\mathbf{k}^i_{t}$
\begin{equation}
\mathbf{k}^i_{t} = \lambda^d(\mathbf{k}^i_{t-1},\mathbf{o}^i_{t-1}; \mathbf W_{\lambda^d})  
\end{equation}
with $t\in[1,T]$ and $\lambda^d$ the decoding LSTM-D layer. The input vector is:

\begin{equation}
	\mathbf{o}^i_{t}=[(\mathbf{h}^i_{t})^T,(\sum_{j\neq i}a^{ij} \mathbf{h}^j_{t} )^T,(\mathbf z)^T]^T
\end{equation} 
\noindent
It stacks information from the encoded history of observations of agent $i$ up to $t$, $\mathbf{h}^i_t$, from the noise vector $\mathbf z$, and from the impact of future trajectories of the neighboring agents $j$, $\sum_{j\neq i}a^{ij} \mathbf{h}^j_t$. The construction of this term is described hereafter. 

\subsubsection{Social Ways: Attention pooling}

The influence of the other agents on agent $i$ is evaluated by encoding the vector $\mathbf X^{\neg i}_{1:T}$, through LSTM-E, and by applying an attention weighting process that produces weights $\mathbf a^i\delequal [a^{i1},..,a^{ij},...,a^{iN}]^T$ for agent $i$. They are defined as in~\cite{SoPhie2018}, for $j\neq i$, based on pre-defined geometric features $\delta^{ij} \in \mathbb{R}^3$ stacking (1) the Euclidean distance between agents $i$ and $j$, (2) the bearing angle of agent $j$ from agent $i$ (i.e. the angle between the velocity vector of agent $i$ and the vector joining agents $i$ and $j$), and (3) the distance of closest approach (i.e. the smallest distance two agents would reach in the future if both maintain their current velocity)~\cite{Kooij2014ContextBasedPP}.

An interaction feature vector between agents $i$ and $j$ is defined as an embedding in $\mathbb{R}^{d_\sigma}$ of the social features $\delta^{ij}$, through a FC layer $\mathbf f^{ij} = \phi(\delta^{ij};\mathbf W_\phi)$. Finally, the attention weights are obtained with the following scalar products and softmax operations between the hidden history vectors $\mathbf h^k$ and the interaction feature vectors $\mathbf f^{ik}$  
{\small
\begin{eqnarray}
\sigma(\mathbf f^{ik},\mathbf h^k) &=& \frac{N-1}{\sqrt{d_\sigma}} <\mathbf f^{ik},\mathbf W_\sigma \mathbf h^k>,\\
a^{ij} &=& \frac{\exp(\sigma(\mathbf f^{ij},\mathbf h^j))}{\sum_{k\neq i}\exp(\sigma(\mathbf f^{ik},\mathbf h^k))}
\end{eqnarray}}
\noindent where $d_\sigma$ is the common number of rows of the embedded features $\mathbf f$ and of the linear mapping $\mathbf W_\sigma$ applied on the hidden features.

\subsubsection{Discriminator}

The Discriminator is described on the right part of Fig.~\ref{fig:blcok_diagram}. It contains two encoding LSTM layers,  one (applied $\tau+1$ times) for observations, and one (applied $T$ times) for predictions, and 2 FC layers to predict the samples labels. It takes as an input either a composite candidate trajectories for agent $i$, $[{\mathbf{x}}^{i}_{-\tau:0},\tilde{\mathbf{x}}^{i,k}_{1: T}]$, or a ground truth trajectory, $[{\mathbf{x}}^{i}_{-\tau:T}]$, and outputs a probability for any of them to have been taken as a sample from the data.

\subsubsection{Training the GAN}

GAN training is known to be hard, as it may not converge, exhibit vanishing gradients when there is imbalance between the Generator and the Discriminator, or may be subject to mode collapsing, i.e. sampling of synthetic data without diversity. When predicting pedestrian motion, it is critical to avoid mode collapsing, as it could result in catastrophic decisions, i.e. for an autonomous driving agent.

Here, we have introduced two major changes in the GAN training. First, we do not use, as in other stochastic prediction methods~\cite{SocialGAN2018, SoPhie2018}, an L2 loss term $\|G(\mathbf{z}|\mathbf{x}^i_{-\tau: 0},\mathbf{X}^{\neg i}_{-\tau: 0})-\mathbf{x}^i_{-\tau:T}\|^2$ enforcing the generated samples to be close to the true data, because we have observed negative impact of this term in the diversity of the generated samples. 

%Also, we have implemented an unrolled version of the GAN training~\cite{Metz2017UnrolledGA}, which, as we will see in the experimental results section, has a very positive impact on avoiding the mode collapsing problem with respect to other versions of GANs.
%
%With the unrolled GAN, the Discriminator $D$ is trained as in a standard version. Being given the Generator parameters $\theta_{G}$, it optimizes the parameters $\theta_D$ through       
%   
%{\small   
%\begin{equation}
%\begin{array}{l}
% \max_{\theta_D} V(\theta_D)= \\
% \;\;\mathbb{E}_{p_{data}(\mathbf{x}^i_{-\tau:T})}[\log D(\mathbf{x}^i_{1:T}|\mathbf{x}^i_{-\tau: 0};\theta_D)]+\\
%\;\;\mathbb{E}_{p_z(\mathbf{z})}[\log(1-D(G(\mathbf{z}|\mathbf{x}^i_{-\tau: 0},\mathbf{X}^{\neg i}_{-\tau: 0};\theta_G);\theta_D))],\label{eq:ganloss}
%\end{array}
%\end{equation}}
%where $\mathbf{z}$ is the noise input. The Generator, however, follows a more elaborate strategy,
%
%{\small
%\begin{equation}
%\begin{array}{lc}
%\min_{\theta_G} V(\theta_G)=&  \\
%\;\;\mathbb{E}_{p_{data}(\mathbf{x}^i_{-\tau:T})}[\log D(\mathbf{x}^i_{1:T}|\mathbf{x}^i_{-\tau: 0};\theta^K_D(\theta_{G},\theta_D))]&+\\
%\;\;\mathbb{E}_{p_z(\mathbf{z})}[\log(1-D(G(\mathbf{z}|\mathbf{x}^i_{-\tau: 0},\mathbf{X}^{\neg i}_{-\tau: 0};\theta_G);\theta^K_D(\theta_{G},\theta_D)))], & \label{eq:ganloss}
%\end{array}
%\end{equation}}
%where the Discriminator applies, given $\theta_G$ and for a current value  $\theta_D$, $K$ consecutive gradient updates and evaluates its gains $K$ steps ahead of the Generator. Hence, it avoids trivial short-term counter-measures from the Generator.

Also, we have implemented an Info-GAN~\cite{Infogan2016} architecture, which, as we will see in the experimental results section, has a very positive impact on avoiding the mode collapsing problem with respect to other versions of GANs. Info-GAN learns disentangled representations of the sources of variation among the data, and does so by introducing a new coding variable $c$ as an input (see Fig.~\ref{fig:blcok_diagram}). The training is performed by adding another term to maximize a lower bound of the mutual information between the distribution of $c$ and the distribution of the generated outputs, which requires training another sub-network $Q(c|\mathbf x_{1:T})$ (with parameters $\theta_Q$) which serves as a surrogate to evaluate the likelihoods $p(c|\mathbf x_{1:T})$ over the generated data $\mathbf x_{1:T}$. The training optimization problem is written as:
	
{\small   
\begin{equation}
\begin{array}{l}
\min_{\theta_G,\theta_Q}\max_{\theta_D} V(\theta_G,\theta_Q,\theta_D)= \\
\;\;\mathbb{E}_{p_{data}(\mathbf{x}^i_{-\tau:T})}[\log D(\mathbf{x}^i_{1:T}|\mathbf{x}^i_{-\tau: 0};\theta_D)]+\\
\;\;\mathbb{E}_{p_z(\mathbf{z})}[\log(1-D(G(\mathbf{z}|\mathbf{x}^i_{-\tau: 0},\mathbf{X}^{\neg i}_{-\tau: 0};\theta_G);\theta_D))]-\\
\;\;\lambda \mathbb{E}_{p(c),p_z(\mathbf{z})}[\log Q(c|G(\mathbf z|\mathbf{x}^i_{-\tau: 0},\mathbf{X}^{\neg i}_{-\tau: 0};\theta_G);\theta_Q)] \label{eq:infoganloss}
\end{array}
\end{equation}}
where $\mathbf{z}$ is the noise input and $c$ the new latent code.

%% file: notations.tex
\def \xObsv {\mathbf{x}^i_{-\tau: 0}}
\def \xPred {\mathbf{x}^i_{1: T}}
\def \XObsv {\mathbf{X}^{-i}_{-\tau: 0}}
\def \XPred {\mathbf{X}^{-i}_{1: T}}

\def \xPredH {\mathbf{\hat{x}}^i_{1: T}}
\def \XPredH {\mathbf{\hat{X}}^{-i}_{1: T}}

\def \TEOM {\mathbf{\hat{m}}^i_{1: T}}

%% file: experiments.tex
% !TEX root = egpaper_for_review.tex

\input{notations}
\subsection{Implementation details}
\label{subsec:implementation}
We implemented our system using PyTorch framework.

First, note that all the internal FC layers of both the Generator and the Discriminator are associated to  LeakyReLU activation functions, with slope $0.1$. 

{\textbf{Generator}: comprises a first FC linear embedding $\mu$ of size $4\times 128$, over positions and velocities. The Encoder block in Generator contains one layer of $128$ LSTM units (LSTM-E). Using 2 continuous latent code, noise vector with length of 62, and pooling vectors of size 64, which totally gives a 256-d vector, the Decoder LSTM (LSTM-D's) then uses 128 LSTM units in one layer and 3 FC layers with size of 64, 32, 2 to decode the predictions.} Weights are shared among LSTM layers with the same function.

\textbf{Discriminator}: uses two LSTM blocks (LSTM-OE and LSTM-PE) with hidden layers of size $128$ to process both the observed trajectories (size $4 \times \tau + 4$) and the predicted/``future'' trajectories (size $4 \times T$); these outputs are processed in parallel with two $64\times 64$ FC layers. Then they are concatenated in fed to two separate FC blocks: soft-classifier (D) $[64 \times 1]$ and latent-code reconstructor $[64 \times 2]$ (Q). Finally, $\tau$ and $T$ are set to $7$ and $12$ respectively.

 In each dataset, we train the GAN network with the following hyper-parameters setting: mini-batch size $64$, learning rate $0.001$ for Generator and $0.0001$ for Discriminator, momentum $0.9$. The GAN is trained for $20000$ epochs.

\subsection{Datasets}

For the evaluation of our approach, we use two publicly
available datasets: ETH~\cite{Pellegrini2009} and UCY~\cite{CrowdsByExample2007}. These datasets consist of real-world human trajectories. They are labeled manually at a rate of 2.5 fps. The ETH dataset contains 2 experiments (coined as ETH and Hotel) and the UCY dataset contains 3 experiments (ZARA01, ZARA02 and Univ).
In order to evaluate the prediction algorithm, each dataset is split into $5$ subsets, where we train and validate our model on $4$ sets and test on the remaining set.

\subsection{Baseline Predictors and Accuracy Metrics}
We consider two sets of baselines.
\begin{enumerate}[leftmargin=*]
\item Deterministic prediction models, that generate one trajectory for each observation:

% \vspace{-4mm}
\begin{itemize}[leftmargin=*]
	\item Linear: This is a simple constant velocity predictor.

	\item S-Force: It uses an energy function based on Social Forces to optimize the next agent action. The function penalizes jerky movements, high minimum distance to other agents and so on. We use the version by Yamaguchi et al. \cite{Yamaguchi2011}, in which a term enforces the agent to stay close to the group it belongs to.
	
	\item S-LSTM~\cite{SocialLSTM2016}: It associates each pedestrian to one LSTM unit (the Social-LSTM) and gathers the hidden states of neighboring pedestrians with a so-called social-pooling mechanism to perform the prediction.

\end{itemize}

\item Stochastic prediction models, that generate a set of samples from a surrogate of the predictive distribution:

\begin{itemize}[leftmargin=*]
	\item Social-GAN: A GAN-based prediction~\cite{SocialGAN2018}. We consider the variants S-GAN-P and S-GAN, with and without a pooling mechanism, respectively.
	
	\item SoPhie \cite{SoPhie2018} which implements Social and Physical attention mechanism in a GAN predictor.
\end{itemize}
\end{enumerate}
Similarly to previous works~\cite{SocialGAN2018, SocialAttention2018Vemula}, we use the following metrics to evaluate the proposed system over the prediction on one testing data $\mathbf{x}^i_{-\tau:T}$:
\begin{enumerate}[leftmargin=*]
\item Average Displacement Error (ADE), averaging Euclidean distances between ground truth and predicted positions over all time steps: 
%\mathbb{E}_{p_{data}(
\vspace{-0.2cm}
{\small\begin{equation}
\mbox{ADE}(\mathbf{x}^i_{-\tau:T})=\frac{1}{T}\sum_{t=1}^T \Vert {\mathbf{x}}^i_t - \mathbf{\hat{x}}^i_t(\mathbf{x}^i_{-\tau,0},\mathbf{X}^{\neg i}_{-\tau,0}) \Vert.
\label{eq_ADE}
\end{equation}}
\item Final Displacement Error (FDE), i.e. Euclidean distance between the ground truth and predicted final position: 
{\small\begin{equation}
\mbox{FDE}(\mathbf{x}^i_{-\tau:T})= \Vert {\mathbf{x}}^i_T - \mathbf{\hat{x}}^i_T(\mathbf{x}^i_{-\tau,0})  \Vert.
\label{eq_FDE}
\end{equation}}
\end{enumerate}

Then, we evaluate the expectations of these errors over all the samples in our testing datasets. We observe $\tau=8$ frames (2.8 seconds) and predict the next $T=12$ frames (4.8 seconds).

To evaluate stochastic models (that generate a set of samples), we use the methodology proposed in~\cite{SocialGAN2018}. We generate $K$ samples and take the closest one to Ground truth for evaluation. Hereafter, we consider $K=20$.

\subsection{Evaluation of Prediction Errors}

\begin{table*}[t]
	\begin{center}
	\begin{tabular}{c||c|c|c||c|c|c|||c}
		
%		\hline		
		\multicolumn{1}{c}{~} &
		\multicolumn{3}{c}{Deterministic Models} &
		\multicolumn{4}{c}{Stochastic Models} \\
		Dataset & Linear & S-Force & S-LSTM & S-GAN & S-GAN-P & SoPhie & S-Ways \\
		\hline \hline
		
		\textbf{ETH}
                & \textbf{0.59} / \textbf{1.22} & 0.67 / 1.52 & 1.09 / 2.35 & 0.68 / 1.26 & 0.77 / 1.38 & 0.70 / 1.43 & \textbf{0.39} / \textbf{0.64} \\

		%\hline
		\textbf{Hotel}
                & \textbf{0.36} / \textbf{0.64} & 0.52 / 1.03 & 0.79 / 1.76 & 0.47 / 1.01 & 0.44 / 0.89 & 0.76 / 1.67  & \textbf{0.39} / \textbf{0.66} \\

		%\hline
		\textbf{Univ}
                & 0.82 / 1.68 & 0.74 / \textbf{1.12} & \textbf{0.67} / 1.40 & 0.56 / \textbf{1.18} & 0.75 / 1.50 & \textbf{0.54} / 1.24 & 0.55 / 1.31 \\

		%\hline
		\textbf{ZARA01}
                & 0.44 / 0.98 & \textbf{0.40} / \textbf{0.60} & 0.47 / 1.00 & 0.34 / 0.69 & 0.35 / 0.69 & \textbf{0.30} / \textbf{0.63} & 0.44 / 0.64 \\

		%\hline
		\textbf{ZARA02}
                & 0.43 / 0.95 & \textbf{0.40} / \textbf{0.68} & 0.56 / 1.17 & \textbf{0.31} / \textbf{0.64} & 0.36 / 0.72 & 0.38 / 0.78 & 0.51 / 0.92  \\
%		\hline
	\end{tabular}
	\end{center}
	\caption{Comparison of prediction error of our proposed method (S-Ways) vs baselines. The ADE and FDE values are separated by slash.}
	%\todo{Add other baselines to the table: S-GAN \& S-GAN-P \& SoPhie \\}
	\label{table:1}
\end{table*}

The average prediction errors for both ADE and FDE metrics are shown in Table~\ref{table:1}. As it can be seen, the use of our approach leads to significantly lower prediction errors for the ETH and Hotel experiments, but not on the ZARA experiments. We attribute this behavior in that, in the ZARA experiments, the width of the waypath for pedestrians is significantly smaller than in the Hotel and ETH scenes. Hence, there is less variance in the trajectories. Our proposed system intrinsically tends to generate various samples that result in good performance with more complex scenes and non-linear trajectories.

Among the deterministic models, though Social-LSTM model uses a much more complex system than its counterparts, it fails to outperform the other baselines and as the authors in \cite{SocialGAN2018} mention it, it needs a synthetic dataset as a second source of training to improve the system accuracy.

\begin{figure*}
\begin{center}
\begin{tabular}{cccc}
	\includegraphics[width=0.22\linewidth]{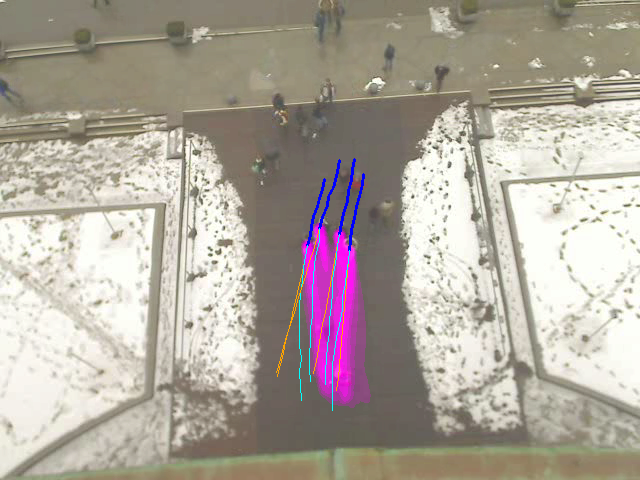}&
	\includegraphics[width=0.22\linewidth]{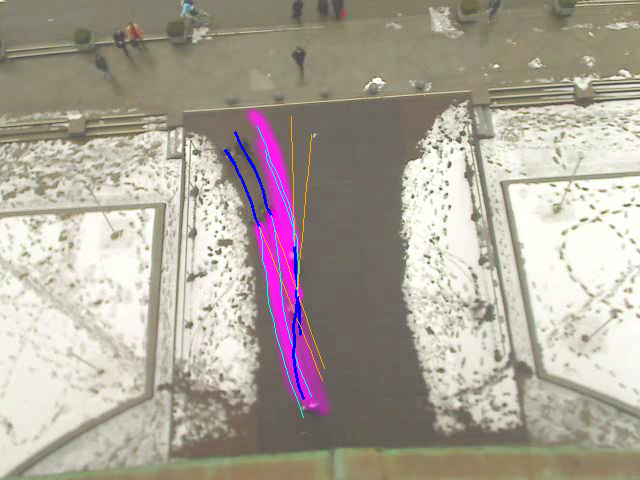}&
	\includegraphics[width=0.22\linewidth]{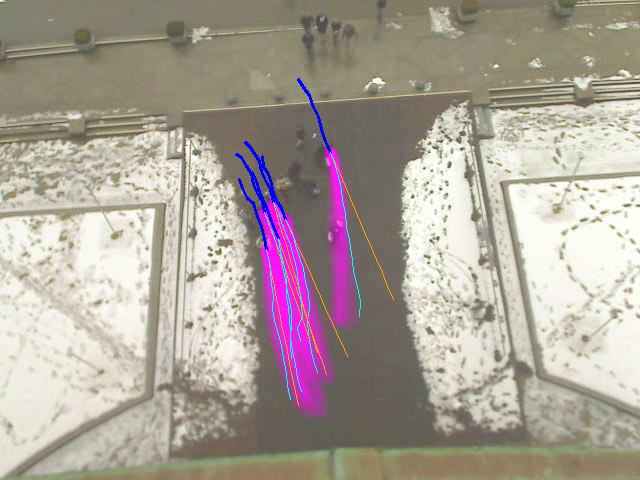}&	
	\includegraphics[width=0.22\linewidth]{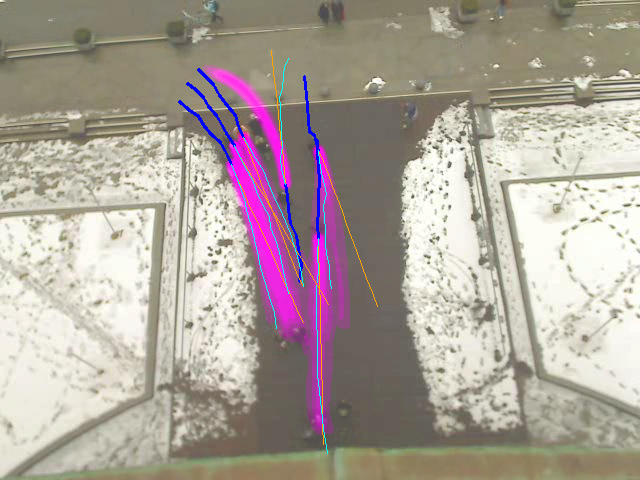}\\
	\includegraphics[width=0.22\linewidth]{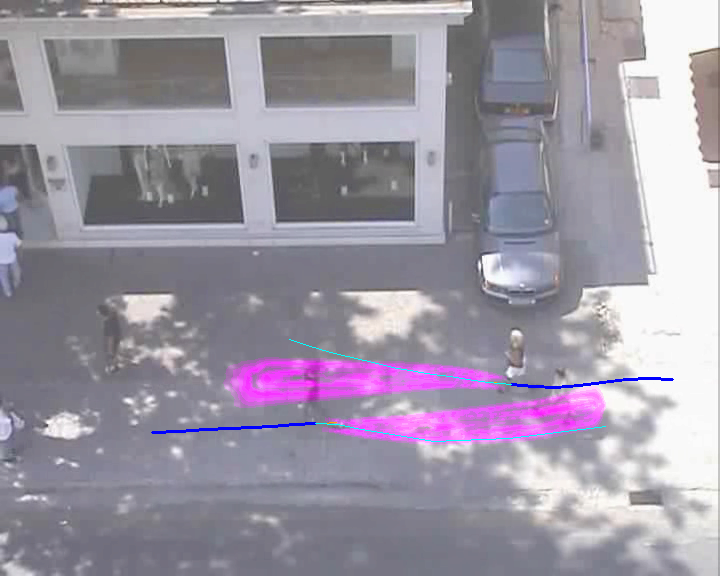}&	
	\includegraphics[width=0.22\linewidth]{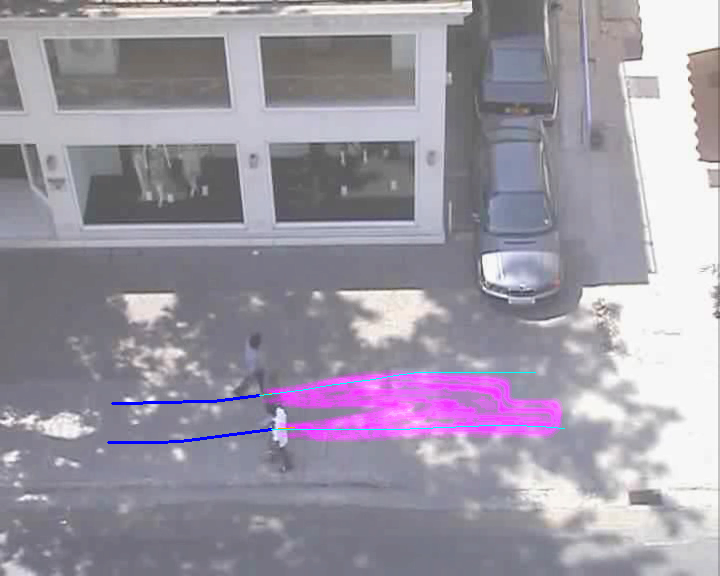}&
	\includegraphics[width=0.22\linewidth]{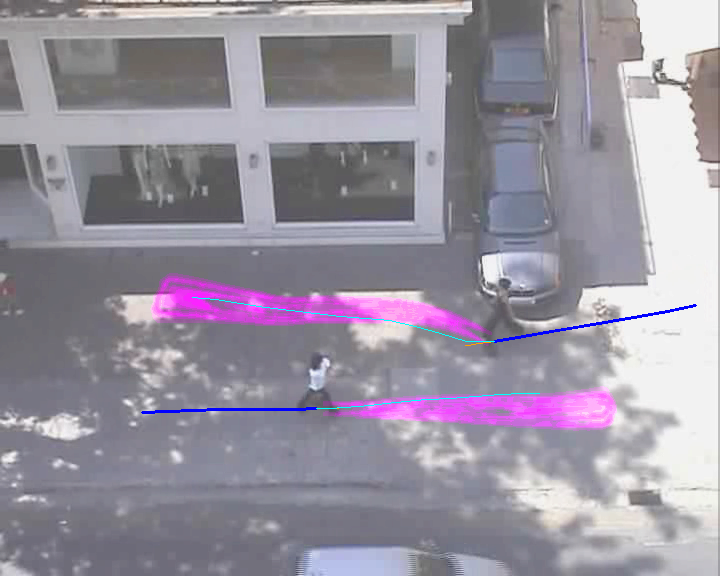}&
	\includegraphics[width=0.22\linewidth]{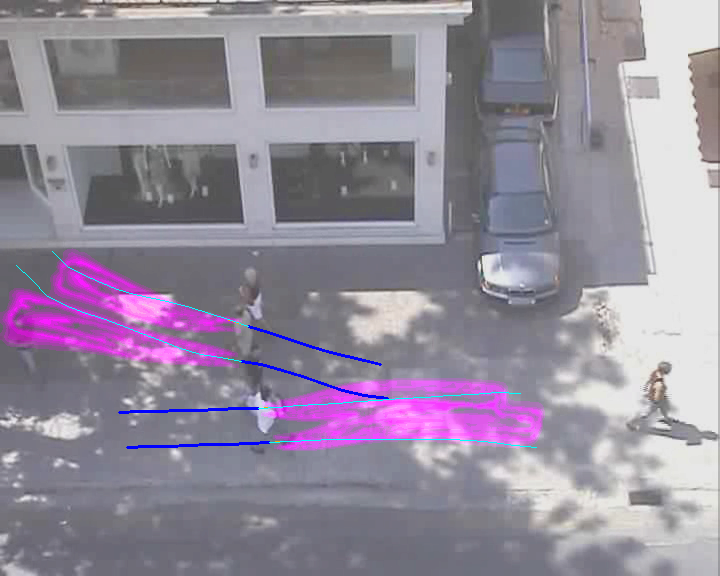}\\
\end{tabular}
\end{center}
\caption{In this figure, we illustrate our sample outputs (in magenta color).
The observed trajectories are shown in blue and ground truth prediction and constant-velocity predictions are shown in cyan and orange lines, respectively. [Best viewed in color.]}
\label{fig:figExample}
\vspace{-0.1cm}
\end{figure*}

In Figure~\ref{fig:figExample}, we give qualitative examples of the outputs and intermediate elements in our approach. We generated 128 samples with our method and the predictive distribution are shown with magenta points. In most of the scenarios (including non-linear actions, collision avoidance and group behaviors), the distribution has a good coverage of the ground truth trajectories and also generates what seems to be plausible alternative trajectories.

%\todo{Qualitative results: We need to show in some frames our predictions are better than S-GAN. And mainly overlap with Ground Truth. Some Failure case can be shown as well.}

%\subsection{Qualitative Evaluation}

\subsection{Quality of the Predictive Distributions}
As commented in Section~\ref{subsec:GAN}, our architecture and its training process are designed to preserve the modes of the predictive trajectory distribution. However, in all the datasets that we have tested, there are very few examples of clearly multi-modal predictive trajectory distributions. Hence, we have created a toy example dataset to study the mode collapsing problem with stochastic predictors.

\begin{figure}[t!]
	\centering
	\includegraphics[clip, trim=1.0cm 0.5cm 1.2cm 1.cm, width=0.6\linewidth]{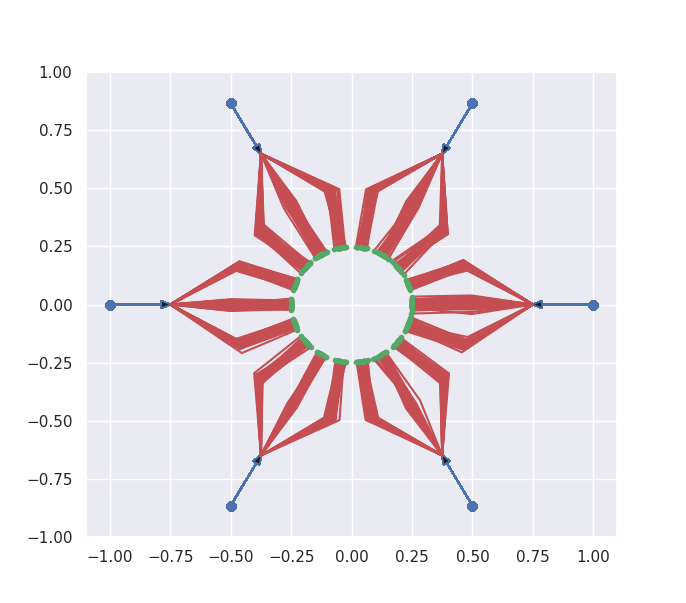}
	\caption{Toy trajectory dataset. There are six groups of trajectories, all starting from one specific point located along a circle (blue dots). When approaching the circle center, they split into 3 subgroups. Their endpoints are the green dots.}
	\vspace{-0.4cm}
	\label{toy_example_fig}	
\end{figure}

This toy example is depicted in Fig. \ref{toy_example_fig}: Given an observed sub-trajectory (blue lines), the Generator should predict the rest of the trajectory (red lines). Each of the 6 groups represents one separate condition to the system ($\mathbf{x}^i_{-\tau:0}$), and each of the 3 sub-groups represents a different mode in the conditional distribution $p(\mathbf{x}^i_{1:T}|\mathbf{x}^i_{-\tau:0})$. Note that the interactions between agents are not considered here.

In order to compare our approach with other GAN-based techniques, we implemented several baselines. In all of them, the prediction architecture is the one we proposed without the attention-pooling; the GAN subsystem changes.

\begin{itemize}[leftmargin=*]
	\item \textbf{Vanilla-GAN}: This is simplest baseline, where the Generator is just trained with the adversarial loss.
	
	\item \textbf{L2-GAN} In addition to adversarial loss, a L2 loss is added to the Generator optimizer.
	
	\item \textbf{S-GAN-V20}: The Variety loss proposed in Social-GAN method \cite{SocialGAN2018} is added to the adversarial loss. This L2-loss only penalizes the closest prediction to ground truth among $V=20$ predictions and gives more freedom to choose prediction samples.

	\item \textbf{Unrolled10}: Vanilla-GAN with the unrolling mechanism proposed in \cite{UnrolledGA2017}. The number of unrolling steps is $10$.

\end{itemize}

\begin{figure}
\begin{center}
\begin{tabular}{cccc}
		\rotatebox{90}{VanillaGAN}
		\includegraphics[width=0.22\linewidth]{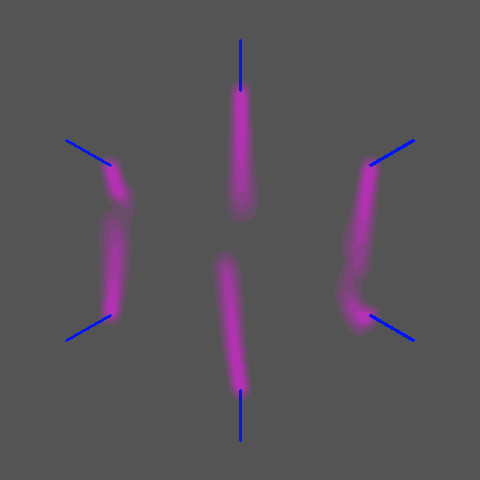}& 
		\hspace{-0.4 cm}
		\includegraphics[width=0.22\linewidth]{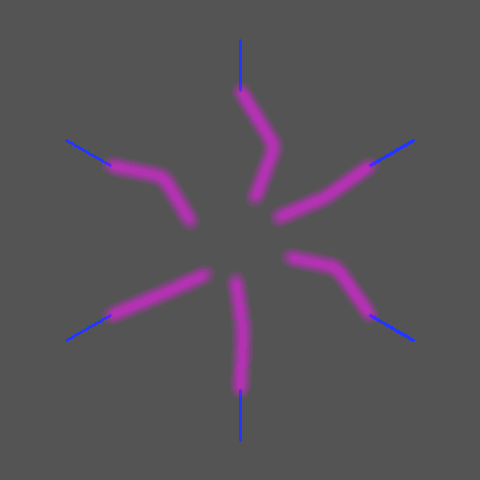}&
		\hspace{-0.4 cm}
		\includegraphics[width=0.22\linewidth]{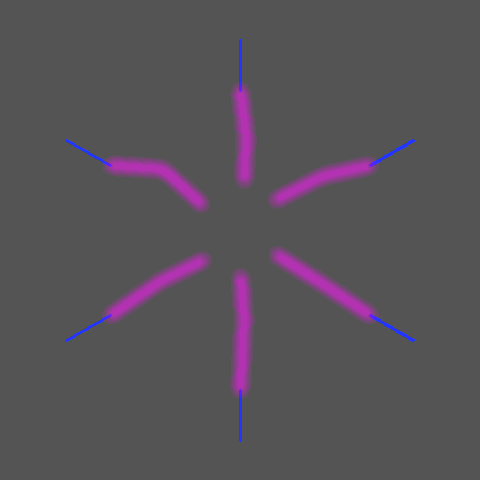}&
		\hspace{-0.4 cm}
		\includegraphics[width=0.22\linewidth]{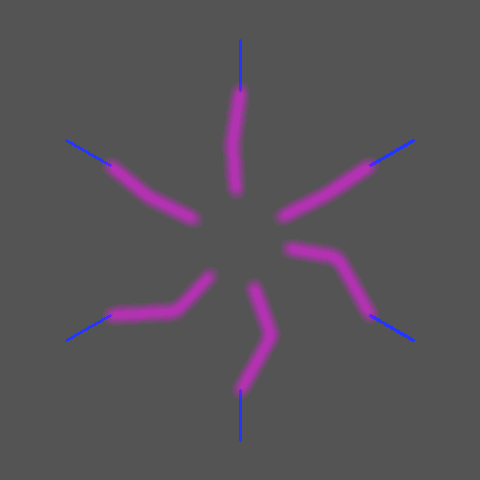} \\

		\rotatebox{90}{~~~L2-GAN}
		\includegraphics[width=0.22\linewidth]{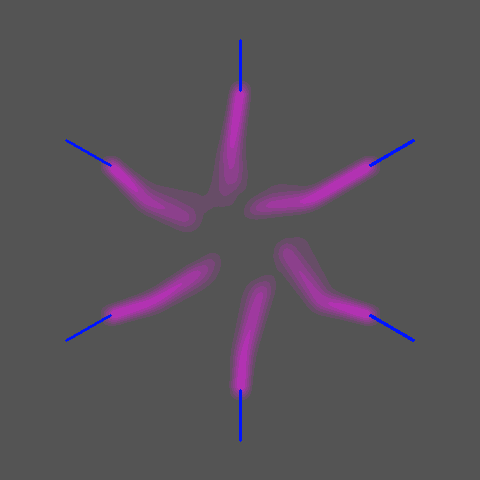}& 
		\hspace{-0.4 cm}
		\includegraphics[width=0.22\linewidth]{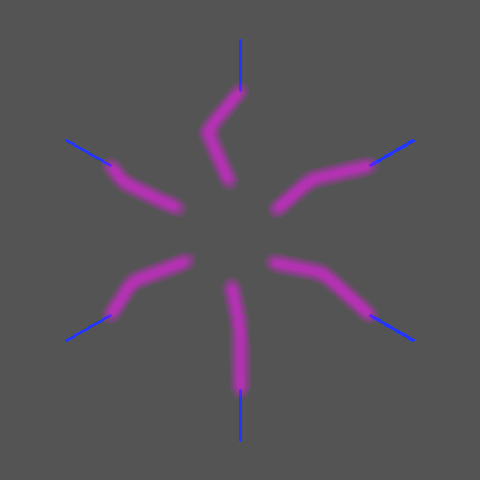}&
		\hspace{-0.4 cm}
		\includegraphics[width=0.22\linewidth]{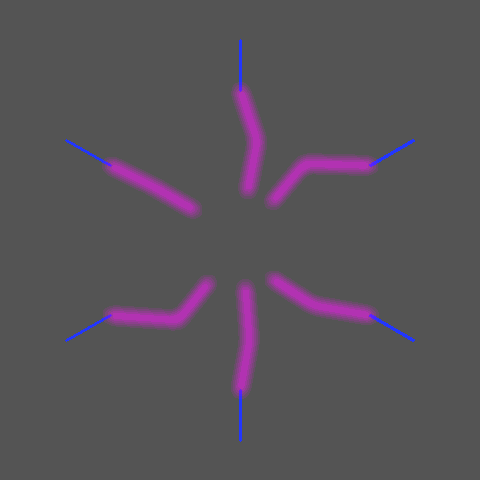}&
		\hspace{-0.4 cm}
		\includegraphics[width=0.22\linewidth]{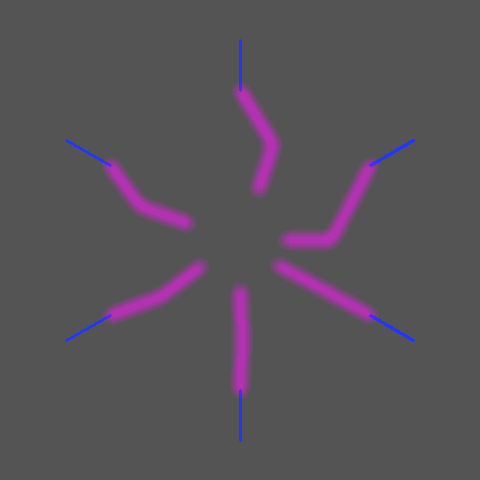} \\

		\rotatebox{90}{S-GAN-V20}
		\includegraphics[width=0.22\linewidth]{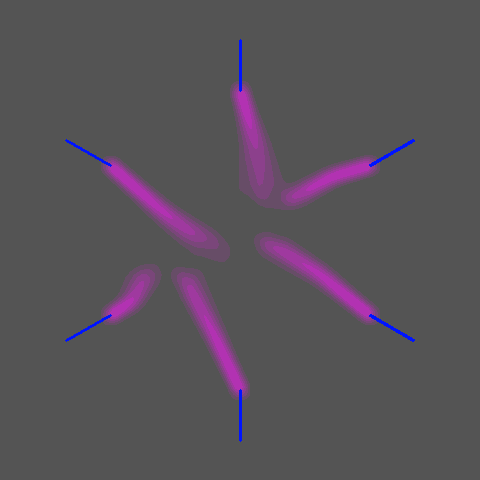}& 
		\hspace{-0.4 cm}
		\includegraphics[width=0.22\linewidth]{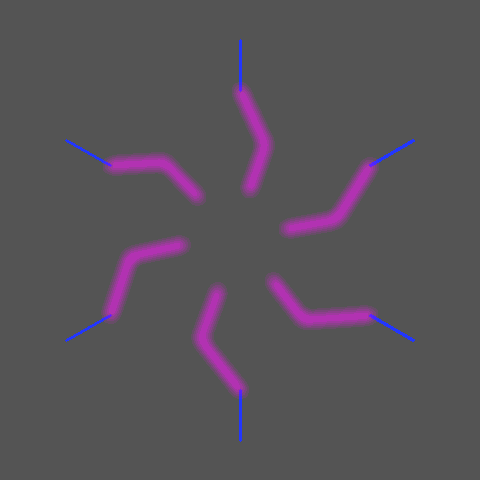}&
		\hspace{-0.4 cm}
		\includegraphics[width=0.22\linewidth]{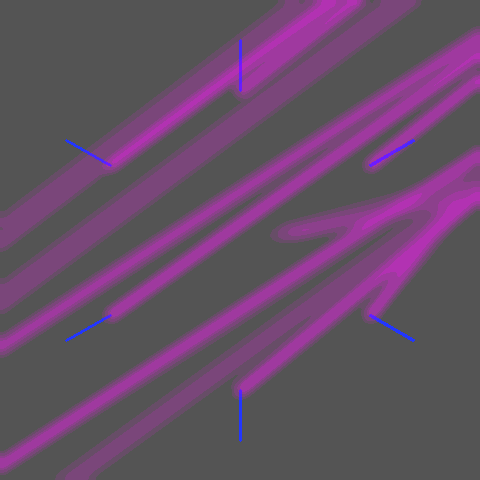}&
		\hspace{-0.4 cm}
		\includegraphics[width=0.22\linewidth]{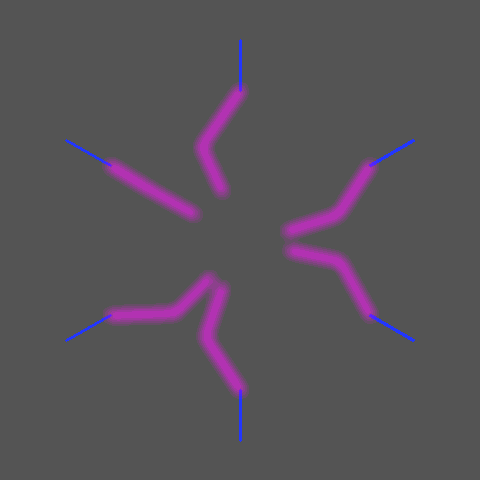} \\

		\rotatebox{90}{~Unrolled10}
		\includegraphics[width=0.22\linewidth]{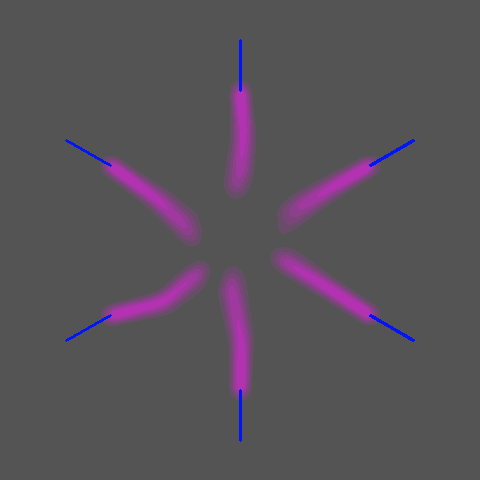}& 
		\hspace{-0.4 cm}
		\includegraphics[width=0.22\linewidth]{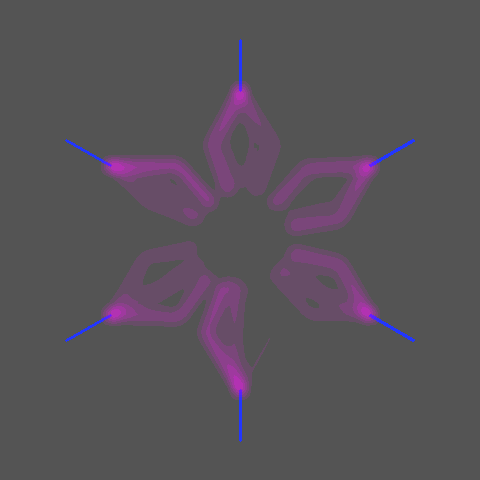}&
		\hspace{-0.4 cm}
		\includegraphics[width=0.22\linewidth]{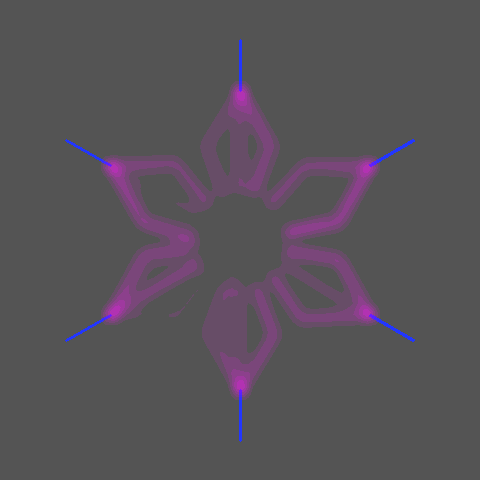}&
		\hspace{-0.4 cm}
		\includegraphics[width=0.22\linewidth]{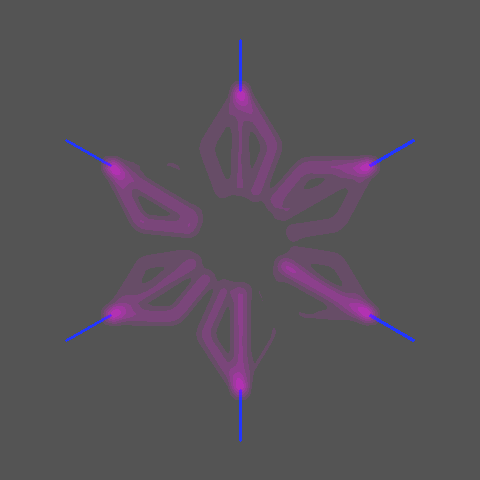} \\

		\rotatebox{90}{~~InfoGAN}
		\includegraphics[width=0.22\linewidth]{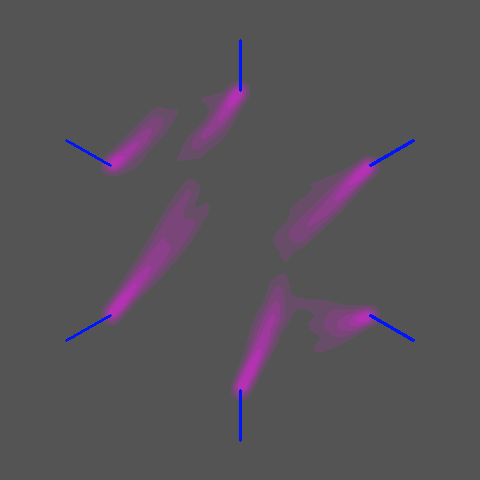}& 
		\hspace{-0.4 cm}
		\includegraphics[width=0.22\linewidth]{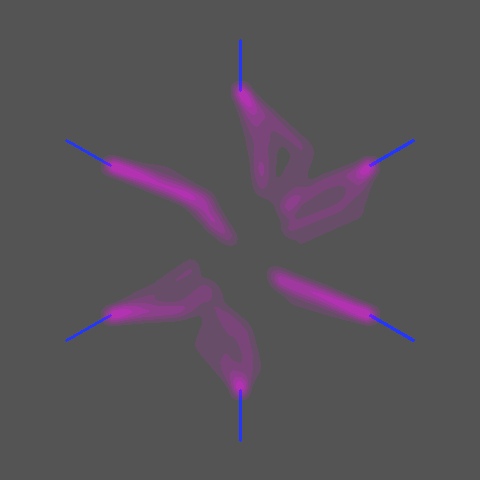}&
		\hspace{-0.4 cm}
		\includegraphics[width=0.22\linewidth]{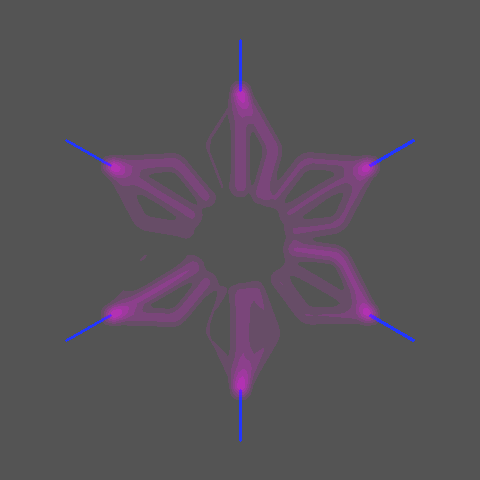}&
		\hspace{-0.4 cm}
		\includegraphics[width=0.22\linewidth]{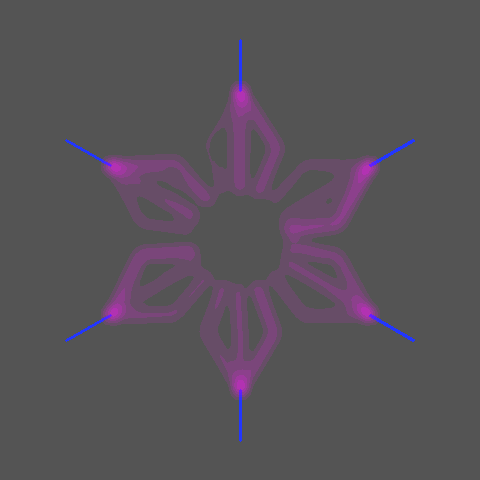} \\	
		Itr=500 & Itr=6000 & Itr=21000 & Itr=90000
\end{tabular}
\end{center}
\caption{Results of learning baselines on Toy Example, for different numbers of iterations. [Best viewed in color.]}
\vspace{-0.2cm}
\label{fig:toy_output}		
\end{figure}

For each of the 6 possible observations, we generate 128 samples, which are depicted in Fig.~\ref{fig:toy_output}. The Info-GAN together with Unrolled-GAN performs the best, with a slight advantage for Info-GAN, since almost all of the modes are preserved successfully after 90,000 iterations. At the same time, Vanilla-GAN, L2-GAN and S-GAN-V20 could not preserve the multi-modality of the predictions. One can see that using L2 loss, the model is converging faster than VanillaGAN and S-GAN-V20.

%\subsubsection{Metrics} 
For a more quantitative evaluation of generative models, we have used the following two metrics to assess the set of fake trajectories versus the set of real samples~\cite{ganMetrics2018Xu}. Given two sets of samples $S_r=\{\mathbf{x}^i_r\}$ and $S_g=\{\mathbf{x}^j_g\}$ with $|S_r| = |S_g|$ and $\mathbf{x}^i_r \sim P_r$ and and $\mathbf{x}^j_g \sim P_g$:
\begin{enumerate}[leftmargin=*]
	\item A 1-Nearest Neighbor classifier, used in two-sample tests to assess whether two distributions are identical. We compute the leave-one-out accuracy of a 1-NN classifier trained on $S_r$ and $S_g$ with positive labels for $S_r$ and negative labels for $S_g$. The classification accuracy for data from an ideal GAN should be close to 50\% when $|S_r| = |S_g|$ is large enough. Values close to 100\% mean that the generated samples are not close to real samples enough. Values close to 0\% mean that the generated samples are exact copies of real samples, and that there is a lack of innovation in such system. 

	\item  The Earth Mover's Distance (EMD) between the two distributions. It is computed as in 
	Eq.~\ref{eq_EMD}:
	{\small
	\begin{equation}
	\begin{split}
	&EMD(P_r, P_g) = \min_{\mathbf w \in \mathbb{R}^{n \times m}}
	\sum_{i=1}^{n} \sum_{j=1}^{m}
	\mathbf w^{ij} d(\mathbf{x}^i_r, ~\mathbf{x}^j_g)		\\
	\mbox{s.t. } & \forall i,j \; \mathbf w^{ij} \geq 0, \sum_{k=1}^m \mathbf w^{ik} = \frac{1}{n} \mbox{ , }\sum_{k=1}^n \mathbf w^{kj} = \frac{1}{m}.
	\end{split}
	\label{eq_EMD}
	\end{equation}}
\end{enumerate}
where $d()$ is called the ground distance. In our case we use the ADE of Eq.~\ref{eq_ADE}, between the future parts of the two trajectories.

\begin{figure*}
	\centering
	\begin{subfigure}
		{\includegraphics[clip, trim=0.8cm 0.6cm 0.0cm 1.0cm, width=0.99\linewidth]{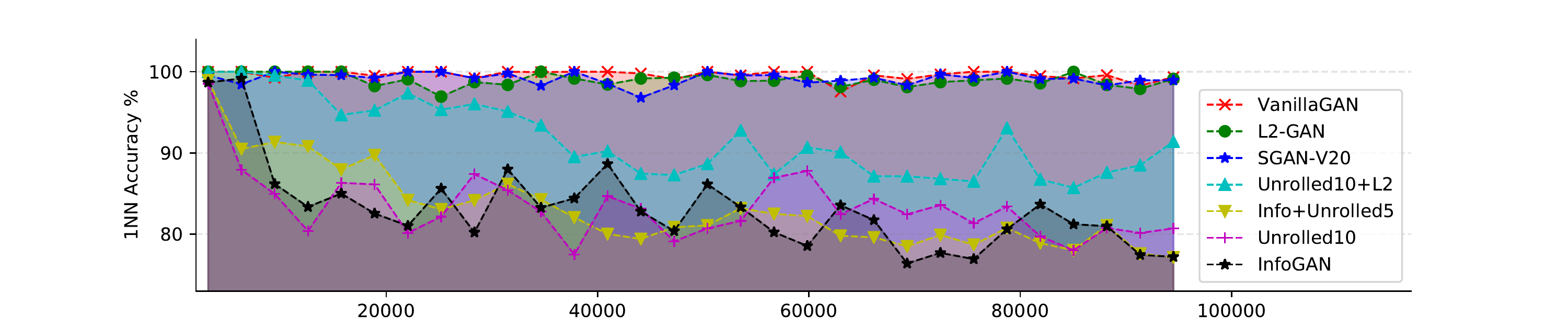}}
		\label{fig:y equals a}
	\end{subfigure}
	
	\begin{subfigure}
		{\includegraphics[clip, trim=0.8cm 0.1cm 0.0cm 1.0cm, width=0.99\linewidth]{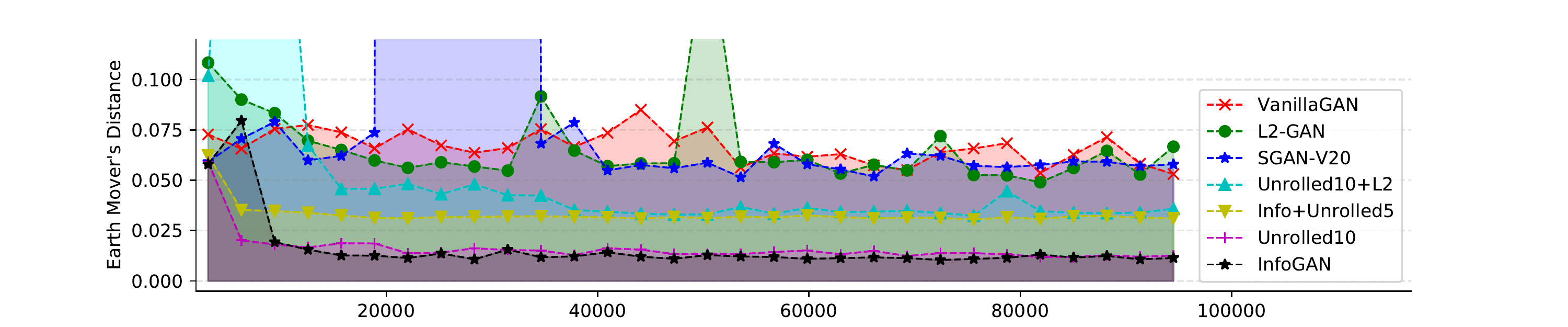}}
		\label{fig:y equals b}
	\end{subfigure}
	\caption{Statistics for different GAN implementations over training iteration. Upper row: 1-NN accuracy metric (closer to \%50 is better). Lower row: Earth Mover's Distance between generated and ground truth samples (the lower, the better).}
	\label{analysis_1nn_emd}	
\end{figure*}

We computed both 1-NN and EMD metrics on our toy dataset with $|S_r| = |S_g| = 20$, for each of the 6 observed trajectories. The results for different baselines are shown in Figures \ref{analysis_1nn_emd}. We added evaluations for a few combinations of the aforementioned baselines (e.g., Info-GAN+unrolling steps or Unrolled+L2).
The lower 1-NN accuracy of our approach using Info-GAN shows its higher performance for matching the target distribution, compared to Vanilla-GAN and other baselines.  It is worth noting that the fluctuations in the accuracies are related to the small size of the set of samples. As it can be seen, Unrolled10 and Info+Unrolled5 have also better performances, while it is obvious that by adding L2 loss, the results are getting worse. The results of the EMD test also proves that both Info-GAN and Unrolled10 offer more stable predictors with lower distances between the fake and real samples. There is no evidence that the Variety loss offers better results than a Vanilla-GAN.

\begin{figure}[!h]
\begin{center}
\begin{tabular}{ccc}
	\rotatebox{90}{Gate-2} &	
	\hspace{-0.2 cm}
	\includegraphics[trim={200 185 200 100},clip, width=0.42\linewidth]{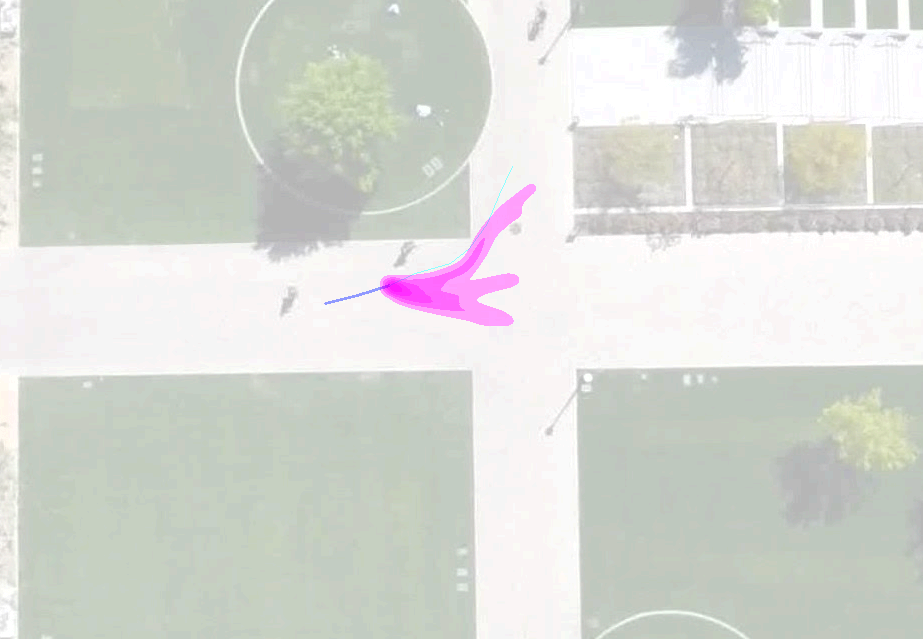} &
	\hspace{-0.2 cm}
	\includegraphics[trim={200 185 205 100},clip, width=0.42\linewidth]{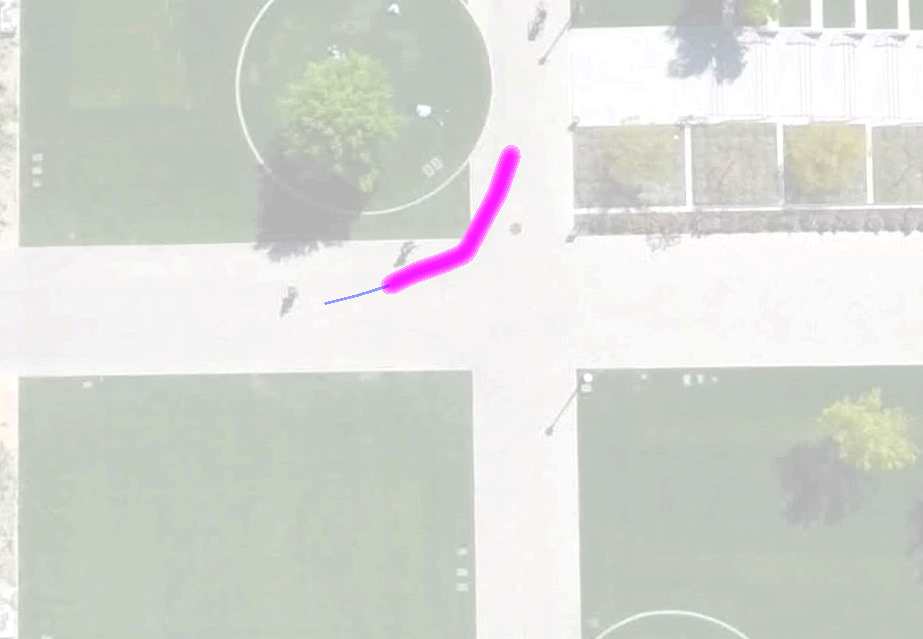} \\
	
	\rotatebox{90}{Hyang-6} &	
	\hspace{-0.2 cm}
	\includegraphics[trim={30 60 555 320},clip, width=0.42\linewidth]{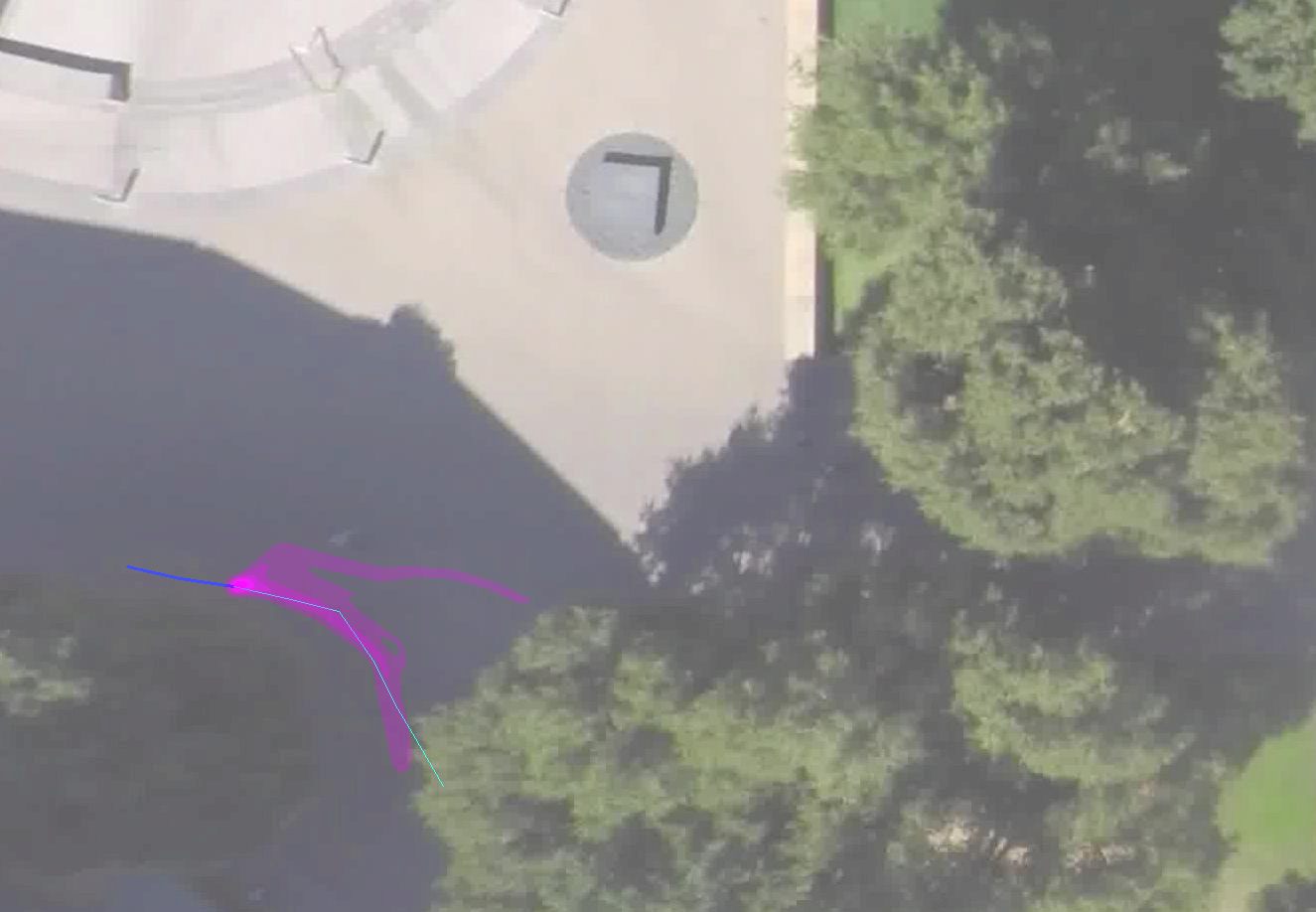} &
	\hspace{-0.2 cm}
	\includegraphics[trim={30 60 555 320},clip, width=0.42\linewidth]{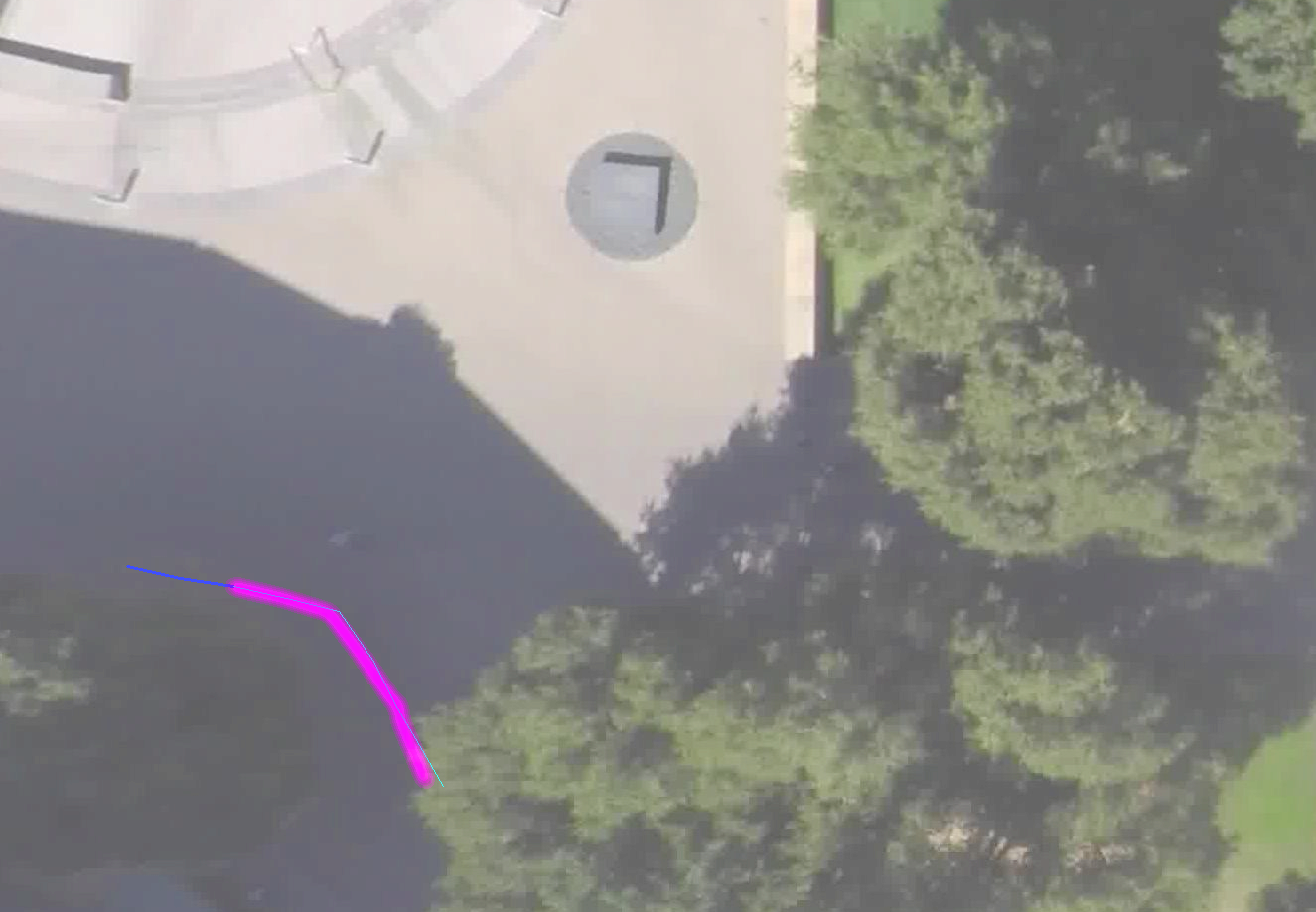} \\
		
	& Social-Ways & Vanilla-GAN
	
\end{tabular}
\end{center}
\caption{Multi-modal trajectory predictive distributions on the SDD dataset: Social-Ways vs. Vanila-GAN. [Best viewed in color.]}
\label{SDD_test}
\end{figure}

Moreover, on real trajectories, we have tested our algorithm on the Stanford Drone Dataset (SDD) \cite{SocialEtiquette2016}. In fact, we have used subsets of trajectories from two scenes (Hyang-6 and Gates-2).
As you see in Fig. \ref{SDD_test}, with our system (left column), separate modes of the predictions appear clearly where the intuition would set them, while the Vanilla-GAN (right column) could not produce various paths.

%\subsection{Effects of GAN improvements}

%\todo{Give some results supporting the claim that the L2-loss is not that necessary}
%\todo{Give some results supporting the use of the unrolled version of GANs}